# Factify 2: A Multimodal Fake News and Satire News Dataset

S Suryavardan*[1], Shreyash Mishra*[1], Parth Patwa[2], Megha Chakraborty[3], Anku Rani[3], Aishwarya Reganti[4], Aman Chadha†[5,6], Amitava Das[3], Amit Sheth[3], Manoj Chinnakotla[7], Asif Ekbal[8] and Srijan Kumar[9]

[1]*IIIT Sri City, India*
[2]*UCLA, USA*
[3]*University of South Carolina, USA*
[4]*Carnegie Mellon University, USA*
[5]*Stanford, USA*
[6]*Amazon AI, USA*
[7]*Microsoft, USA*
[8]*IIT Patna, India*
[9]*Georgia Tech, USA*

Abstract

The internet gives the world an open platform to express their views and share their stories. While this is very valuable, it makes fake news one of our society's most pressing problems. Manual fact checking process is time consuming, which makes it challenging to disprove misleading assertions before they cause significant harm. This is he driving interest in automatic fact or claim verification. Some of the existing datasets aim to support development of automating fact-checking techniques [1, 2], however, most of them are text based. Multi-modal fact verification has received relatively scant attention. In this paper, we provide a multi-modal fact-checking dataset called FACTIFY 2, improving Factify 1 by using new data sources and adding satire articles. Factify 2 has 50,000 new data instances. Similar to FACTIFY 1.0, we have three broad categories - support, no-evidence, and refute, with sub-categories based on the entailment of visual and textual data. We also provide a BERT and Vison Transformer based baseline, which achieves 65% F1 score in the test set. The baseline codes and the dataset will be made available at https://github.com/surya1701/Factify-2.0.

## 1. Introduction

With social media platforms taking center stage as news mediums, shifting facts from fake news has become a cause for concern. Fake news articles typically manifest as fabricated stories

*Equal contribution.

†Work does not relate to position at Amazon.

suryavardan.s19@iiits.in (S. Suryavardan*); shreyash.m19@iiits.in (S. Mishra*); amitava@mailbox.sc.edu (A. Das)

with no verifiable facts, sources, or quotes. Sometimes these stories may be propaganda that is intentionally designed to mislead the reader or may be designed as "clickbait" written for economic incentives. The technological ease of copying, pasting, clicking, and sharing content online has helped misinformation and disinformation to proliferate. This has causes several challenges in events like Covid-19 [3, 4, 5], elections [6] etc. In some cases, stories are designed to provoke an emotional response and placed on certain sites to entice readers into sharing them widely. In other cases, "fake news" articles may be generated and disseminated by "bots" - computer algorithms that are designed to act like people sharing information, but can do so quickly and automatically [7]. Although there are a few large-scale efforts to identify fake news, like FEVER [1] and LIAR[2], these datasets do not account for the evolution of fake news in the real world. Another hindrance to fake news detection on social media platforms is the fact that online information is very diverse, covering a large number of subjects, which contributes complexity of this task. Often times, the truth and intent of any statement are challenging to be verified by computers alone, so efforts must depend on collaboration between humans and technology, à la human-in-the-loop setting [8]. Additionally, the visual cues that support textual claims would help the system to detect fake content with greater confidence. These concerns were addressed in the previous iteration - FACTIFY 1, which released a multimodal fact-checking dataset for multimodal fact verification. The dataset contains images, textual claims, and reference textual documents/images. It proposed a multimodal entailment task to tag these claims against the verified document/image using 3 classes, i.e., support, no-evidence, and refute; each of these categories is explained in the next section. The first two categories are further sub-divided into text and multimodal components. Thus, in total, all the data samples are labeled with one out of five choices. The data was obtained from twitter handles of popular news channels from two large nations – the US and India. Factify 2 is the latest iteration of factify, where we release new data of 50k instances including satirical articles, which utilize a different manner of presentation of fake news.

The paper is organized as follows: Related work is described in section 2. The proposed task is described in section 3. Data collection and data distribution are explained in section 4 while section 5 demonstrates the baseline model. Section 6 shows the results of our baseline models. Finally, we summarise our task along with the further scope and open-ended pointers in section 7.

## 2. Related Work

**Text based dataset:** In recent years, a number of textual datasets for fact-checking and fact-verification have been released. The LIAR [2] dataset contains 13k statements from politiFact [9] annotated into 6 fine-grained labels. FEVER provides manually updated 185k instances of Wikipedia claims and associated supporting documents, categorised as `Support`, `Refute`, or `NotEnoughInfo`. Patwa et al. [10] released a dataset of 10k tweets/articles on Covid-19 annotated as true or false. A dataset for evidence extraction, document retrieval, stance detection, and claim validation is proposed in [11]. [12] create a dataset to differentiate fake news from satire. The PUBHEALTH [13] data has 12k public health claims along with explanations by journalists to support the fact-check labels. Other datasets include [14, 15, 16, 17]. Common methods to detect text based fake news involve the use of CNN [18], RNN [19, 20], BERT [21, 22, 23], etc.

**Multimodal datasets:** Text-only databases are inadequate in the social media era. It is crucial to go beyond and consider additional modalities like image and video to detect fake news. The fakeddit [24] dataset contains one million text+image instances taken from reddit and labeled into 6 fine-grained classes for fake news detection. FakeNewsNet provides spatio-temporal and visual data along with news and social context for analysing and detecting fake news. It contains twitter user data such as location, replies, retweets, timestamps, etc. for about 20k multimodal articles from PolitiFact and GossipCop. A multimodal fact-checking dataset called MOCHEG [25] consists of 21,184 assertions, each of which is given a veracity label (`support`, `refute`, and `not enough information`) and an explanation statement. A video dataset consisting 180 verfied and 200 debunked videos is provided by [26]. Some other datasets are [27, 28, 29].

Modelling approaches to this task are varied and unique in their use of classifiers, adversarial training, attention, etc. SpotFake [30] derives textual and visual representations from BERT and VGG, respectively, before concatenating them for classification. EANN [31] trains a fake news classifier adversarially by adding a event discriminator that ensures that the input data is event-invariant such that newly emerging events can also be verified. CARM-N [32] proposes a multichannel nonvolutional neural network that can mitigate the influence of noise information which may be generated by crossmodal attention fusion by extracting textual feature representation from original data and fused textual information simultaneously. Other methods include use of BERT-based CapsNet [33], Cross-modal similarity [34] and Variational Autoencoders [35] among others [36, 37, 38, 39].

**Factify 1:** FACTIFY [40], is one of the largest multimodal fact-verification public datasets, which includes 50k data points and covers news from India and the US. Images, texts, and reference texts are all part of FACTIFY. They are categorised into three primary groups: `Support`, `Insufficient`, and `Refute`, with additional groups dependent on the inclusion of visual and textual data. FACTIFY 2 follows a similar pattern and releases additional 50k instances which incorporate data from satirical articles and new data sources.

For factify 1, researchers used methods like BERT [41], RoBERTa [42], and BigBird [43] for textual features and ResNet [43], DeiT [44], EfficientNet [45], and VGG [42] for visual features. Please refer to [46] for details of all the methods.

## 3. The Factify Task

Fact verification is a difficult task to completely automate, especially in the case of multimodal data, given the inherent challenges in doing a holistic evaluation with both the vision and text modalities to ascertain the veracity of the claim. To this end, we model fake news verification as a multimodal entailment task such that the veracity of both the text and image is verified.

The formulation of the task is similar to the previously presented Factify 1. Each sample contains a claim that has to be verified or fact checked. Each claim is accompanied by a supporting document that is to be used to determine the veracity through a comparison or entailment based approach. The claim and document are multi-modal i.e. they have textual and visual data enabling multi-modal entailment for fact verification. Each sample has pairs of text, image and OCR.

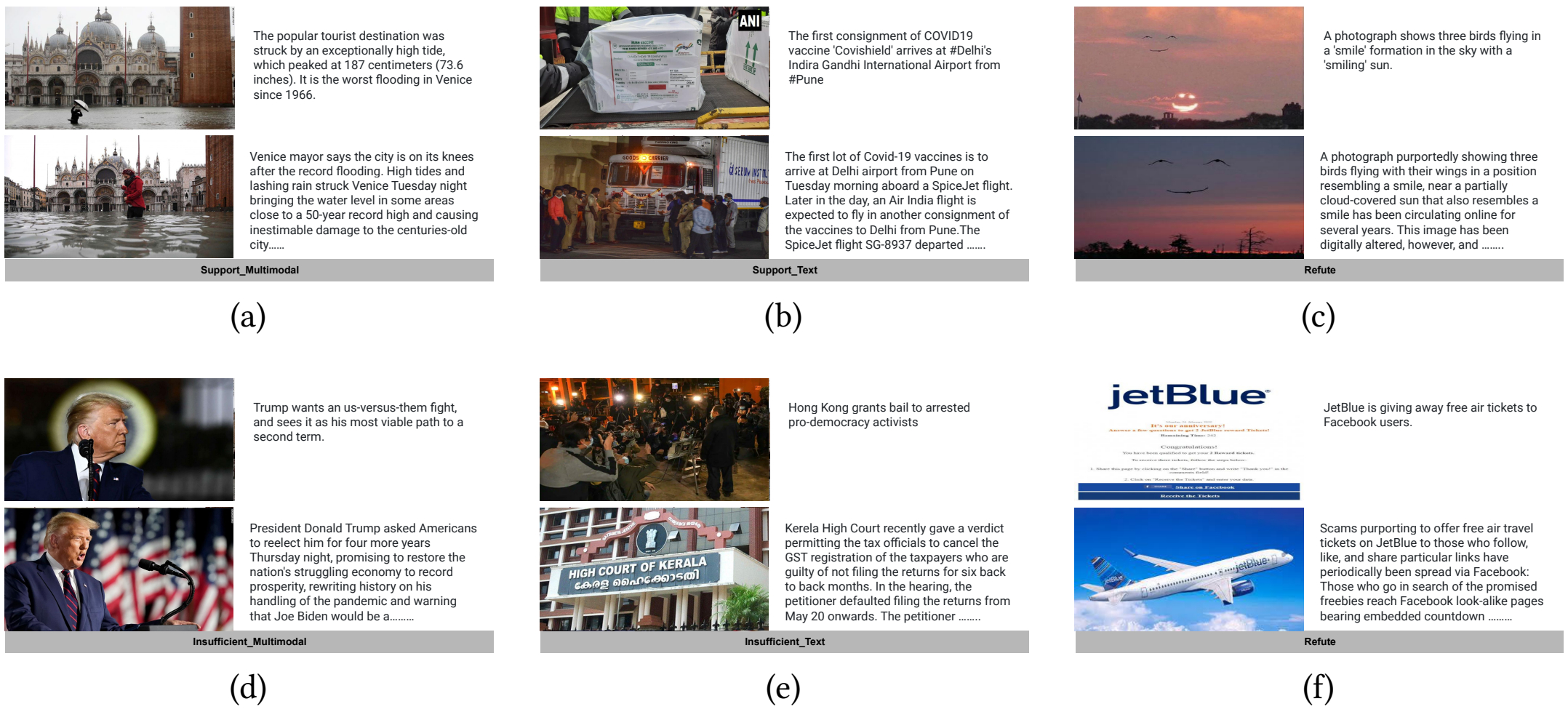

**Figure 1:** These are examples for all the 5 categories. The document text supports the claim text in images (a) and (b), it is insufficient in images (d) and (e), while it refutes the claim in images (c) and f). The claim and document images are entailed in images (a) and (d) and not entailed in images (b) and (e).

We define the following five categories to describe the entailment of the claim and document: `Support_Text`, `Support_Multimodal`, `Insufficient_Text`, `Insufficient_Multimodal`, and `Refute`. The specific description of these categories is as follows:

- `Support_Text`: the textual data for the claim and document are entailed but their images are not entailed.
- `Support_Multimodal`: the textual data is entailed and the images are also similar for the claim and document.
- `Insufficient_Text`: the textual data is not entailed but the claim and document may have several common words, and the images are not entailed.
- `Insufficient_Multimodal`: the claim and document text are not entailed but they may have common words and the images are also entailed in this case.
- `Refute`: The document text and image both contradict or refute the claim text and image, thus, indicating that the given claim is false.

Some examples from the dataset are given in Figure 1.

# 4. Data

In this section, we describe the data collection and data analysis.

## 4.1. Data Collection

The collection process includes two separate pipelines: (i) to collect real news articles for support and no-evidence classes, and (ii) to collect fake news articles for no-evidence and refute classes. The end goal was to curate a dataset with text and image for both claims and their corresponding supporting documents.

The first part of the collection was similar to FACTIFY 1. We collected tweets date-wise from renowned twitter news handles, namely Hindustan Times, ANI and ABC, CNN for India and USA, respectively. The nature and format of these handles as well as their tweets aided our objective of collecting real news claims and articles. To improve the diversity and functionality of the dataset, we compared tweets across the news handles to identify tweets that were reporting the same or similar news. For this, we followed steps similar to FACTIFY 1, where we compare tweet texts using Sentence BERT [47] and using a threshold we categorise whether it is the same news or not. Specifically, we use the pre-trained `paraphrase-MiniLM-L6-v2` [48] variant of Sentence BERT (SBERT) [49] instead of alternatives such as BERT or RoBERTa, owing to its rich sentence embeddings yielding superior performance [47], while being much more time-efficient. If the news is not the same, we compare common words using the NLTK library [50] to categorise the tweets as similar or dissimilar. This helps define the support and no-evidence category respectively as described in Section 3. The similarity between images in the compared tweet pairs are also used to further categorise the data based on visual entailment. Thresholds were set for image similarity to categorise them as entailed or not, based on two metrics: cosine similarity between ResNet50 embeddings and Histogram similarity. With this collected data, we treated the tweet from one handle as the claim and the news article associated with the tweet from the other handle as the supporting document.

The second part is the collection from several different websites. A part of the data for refute category was collected from fact checking websites, similar to FACTIFY 1. We scraped data from Snopes [51], Factly [52] and Boom [53]. These websites provided a well-defined claim and a document disproving the given claim. We added an additional data source in this iteration of the task, we collected satirical articles that were fake in nature but were written in a way that seems real to the reader. While the websites we scraped from i.e. Fauxy [54] and EmpireNews [55], specify that their articles are not true, we added them to the support category. This is because, as aforementioned, the articles support their claim despite the claim being fake in nature. To make the claim multi-modal, we scraped images by searching for the headline of the article. We also manually annotated some articles we collected from the search results of these headlines to add data to the no-evidence and refute category in cases where the articles were about these satirical claims.

### 4.2. Data Statistics And Analysis

The second iteration of FACTIFY has the same categories as FACTIFY 1, with 50,000 data samples. The samples are equally divided among all five categories with a split of 70:15:15 into train, validation, and test sets respectively.

Key words can be vital when identifying or predicting the veracity of a given claim. By analyzing the claim and their documents, we find the most frequently occurring words in Figure 2. Most of the words relate to politics, indicating the bias in the news articles.

The political inclination of the dataset is re-iterated by the word cloud for the support and no-evidence category in Figure 3. However, in the same image, the refute category has a more general distribution of words, with several words related to social media present in the refute category word cloud. We further present unique n-gram examples for the FACTIFY 2 dataset in Table 2 to show the lexical diversity of the dataset.

| | Train | Validation | Test | Total |
|---|---|---|---|---|
| Support_Multimodal | 7000 | 1500 | 1500 | 10000 |
| Support_Text | 7000 | 1500 | 1500 | 10000 |
| Insufficient_Multimodal | 7000 | 1500 | 1500 | 10000 |
| Insufficient_Text | 7000 | 1500 | 1500 | 10000 |
| Refute | 7000 | 1500 | 1500 | 10000 |
| Total | 35000 | 7500 | 7500 | **50000** |

**Table 1**
Dataset distribution statistics for the FACTIFY 2 dataset. Note that the data is balanced across categories.

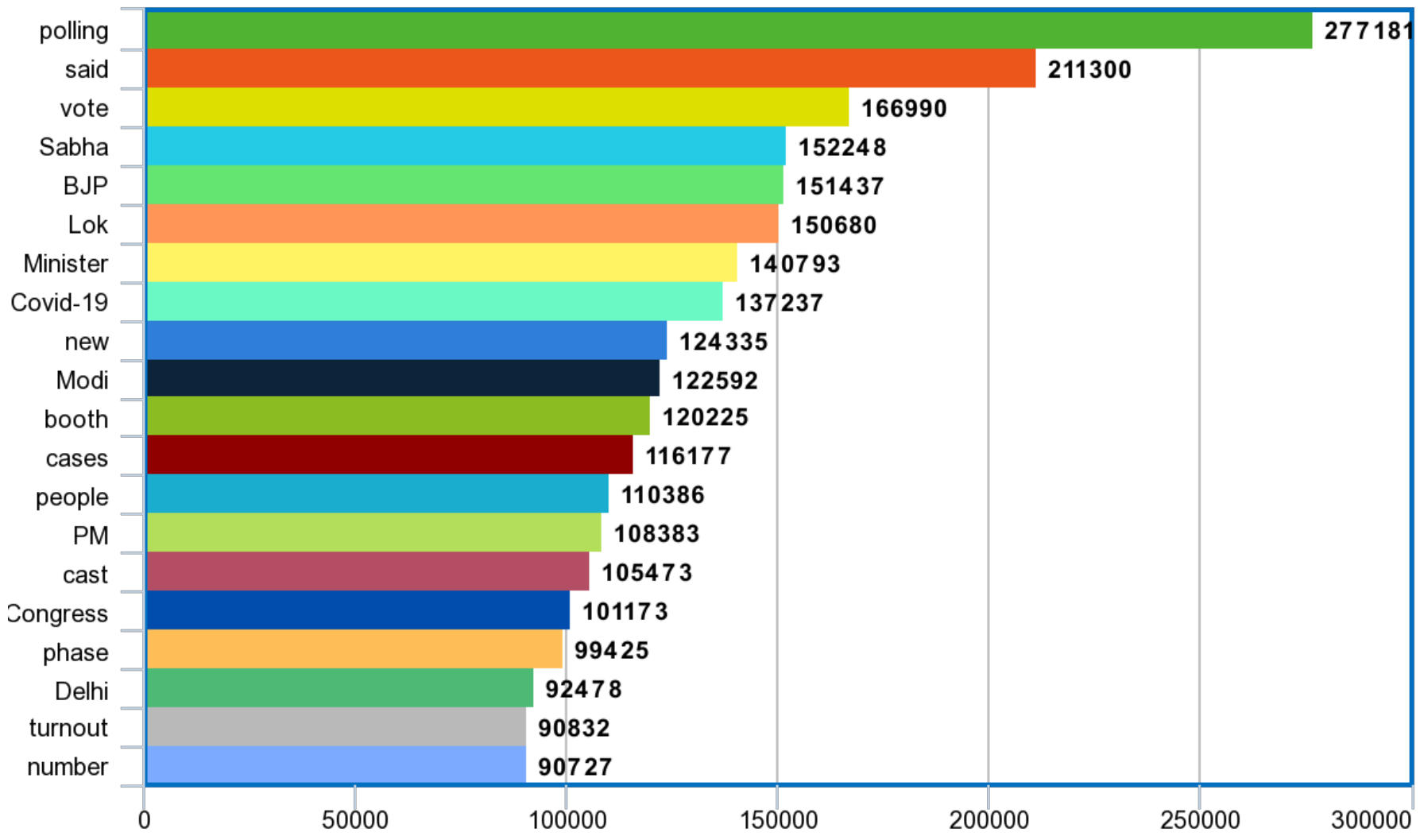

**Figure 2:** The top 20 most frequent words extracted from the claim documents and their frequencies. Many of the words are related to politics.

| N-gram | Examples |
|---|---|
| 1-gram | (polling), (vote) |
| 2-gram | (lok,sabha), (polling,booth) |
| 3-gram | (Prime,Minister,Narendra), (access,to,newsletters) |
| 4-gram | (Prime,Minister,Narendra,Modi), (phase,of,Lok,Sabha) |

**Table 2**
N-gram examples for the claims of all categories.

## 5. Baseline model

Several media are regularly used in online information exchange. Pictures have the power to misrepresent a claim and propagate erroneous information. We must consider both the image and the text in order to appropriately classify the claims. Features must be obtained from claim and document image-text pairings because it is an entailment-based technique. The visual features are obtained from the pre-trained Vision transformer model (ViT) [56]. Thanks to the positional embedding of picture patches carried out by ViT, the ViT model can surpass conventional CNNs in terms of computation and accuracy. Using a pretrained Sentence BERT model (specifically, the `stsb-mpnet-base-v2` variant), the model generates sentence embeddings of claim and document attributes. The Sentence-BERT embedding is concatenated with the pooled output from the ViT model. After passing through an MLP, the combined features are then categorised. The multi-modal characteristics are employed for all three of the sub-tasks after modifications to the MLP. The model architecure is displayed in Figure 4. The codes will be made available at https://github.com/surya1701/Factify-2.0.

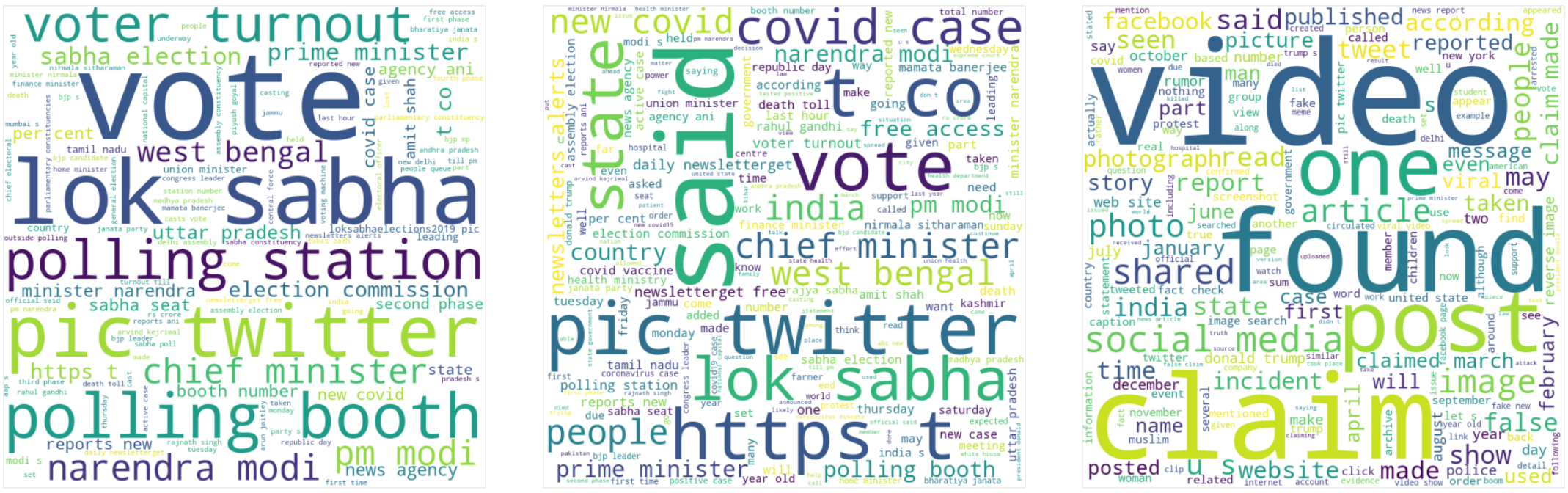

(a) Support (b) Insufficient (c) Refute

**Figure 3:** Word clouds indicating top words used in each class. Words related to politics and Covid dominate in the support and insufficient categories.

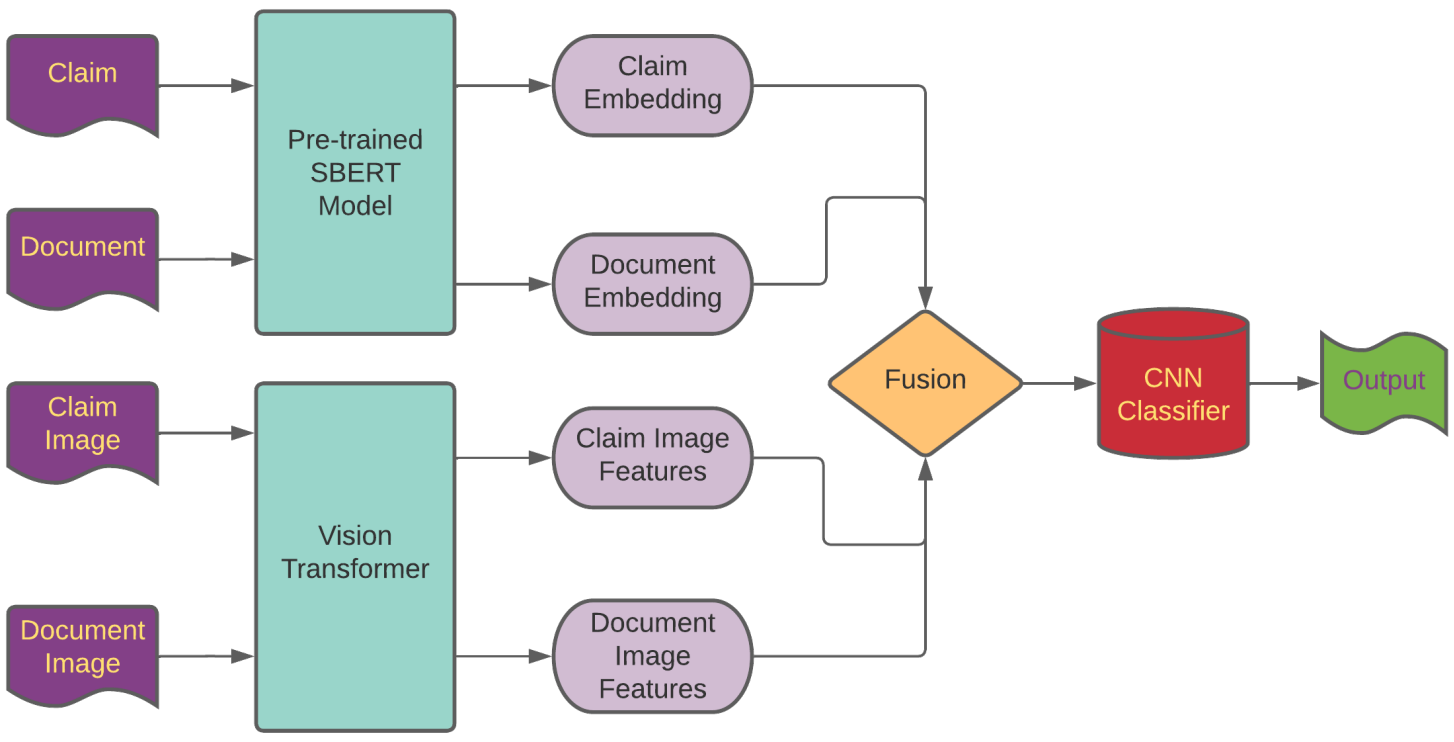

**Figure 4:** Baseline model architecture. Text, image features extracted from the document and the claim are concatenated and used for final prediction.

## 6. Results

Baseline results in Table 3 show Macro F1 scores for some muti-modal modelling approaches mentioned below. Using ViT for extracting visual features and Sentence-BERT for the textual features, the baseline model scores 0.6499. We also compare the baseline model (ViT + SBERT-MPNet) with other methods, such as ViT + SBERT-RoBERTa, in which the SBERT-RoBERTa model is used in place of SBERT-RoBERTa for generating text embeddings. For the Resnet50 + SBERT-RoBERTa and Resnet50 + SBERT-MPNet, a simple ResNet50 model is used to extract visual features. The improvement on using the Vision transformer over the ResNet model signifies the importance of images for the task.

| Method | Macro F1 |
|---|---|
| Resnet50 + SBERT-RoBERTa | 0.4504 |
| Resnet50 + SBERT-MPNet | 0.4727 |
| ViT + SBERT-RoBERTa | 0.6226 |
| **ViT + SBERT-MPNet** | **0.6499** |

**Table 3**
Baseline scores on the test set. ViT based models significantly outperform resnet based models.

## 7. Conclusion and Future Work

By publishing a sizable real-world dataset containing inputs from two modalities, namely text and image, we make a significant step towards creating machine learning approaches for the multimodal fact verification in this study. To underline the difficulties of the issue and the scope for improvement, we conduct data analysis and release multimodal baselines. However, there are a lot of additional research possibilities that can be explored since our work merely touches the surface. One potential research direction could be to enrich the dataset with reasoning that why is a particular news fake. Another possibility is to use synthetic data that matches the general data distribution, thus adding complexity to the refute category.